\documentclass[runningheads]{llncs}
\usepackage{graphicx}
\usepackage{multirow}
\usepackage{caption}
\usepackage{subcaption}
\usepackage{authblk}
\DeclareUnicodeCharacter{0301}{\'a}
\begin{document}
\title{Improved Bengali Image Captioning via deep convolutional neural network based encoder-decoder model}

\def\correspondingauthor{\footnote{Corresponding author: saiful-cse@sust.edu}}
\author{Mohammad Faiyaz Khan \inst{1}\orcidID{0000-0002-2155-5991}
\and
 S.M. Sadiq-Ur-Rahman\inst{2}\orcidID{0000-0003-2428-6595} \and
Md. Saiful Islam\inst{3}\orcidID{0000-0001-9236-380X}\correspondingauthor{}}
\authorrunning{M. Faiyaz Khan et al.}

\institute{Department of Computer Science and Engineering,\\ Shahjalal University of Science and Technology, Sylhet, Bangladesh
\email{mfaiyazkhan@student.sust.edu}\and
\email{shifathrahman472533@gmail.com}\and
\email{saiful-cse@sust.edu}}

\maketitle
\begin{abstract}
Image Captioning is an arduous task of producing syntactically and semantically correct textual descriptions of an image in natural language with context related to the image. Existing notable pieces of research in Bengali Image Captioning (BIC) are based on encoder-decoder architecture. This paper presents an end-to-end image captioning system utilizing a multimodal architecture by combining a one-dimensional convolutional neural network (CNN) to encode sequence information with a pre-trained ResNet-50 model image encoder for extracting region-based visual features. We investigate our approach's performance on the BanglaLekhaImageCaptions dataset using the existing evaluation metrics and perform a human evaluation for qualitative analysis. Experiments show that our approach's language encoder captures the fine-grained information in the caption, and combined with the image features, it generates accurate and diversified caption. Our work outperforms all the existing BIC works and achieves a new state-of-the-art (SOTA) performance by scoring 0.651 on BLUE-1, 0.572 on CIDEr, 0.297 on METEOR, 0.434 on ROUGE, and 0.357 on SPICE.  

\end{abstract}

\section{Introduction}
Automatic image captioning is the process of generating a human-like description of an image in natural language. It is a significantly challenging task as it requires identifying salient objects in the image, understanding their relationships, and generating relevant descriptions of these image features in natural language. The process of generating captions of the images can be applied to automate self-driving cars, implement facial recognition systems, aid visually impaired people, describe CCTV footage, improve image search quality, etc. Despite numerous research attentions in encoder-decoder-based image captioning in the English language, a few works are done in BIC. Researches in these languages can have far-reaching consequences in solving many region-based socio-economic problems.

\begin{figure}
\includegraphics[width=\textwidth]{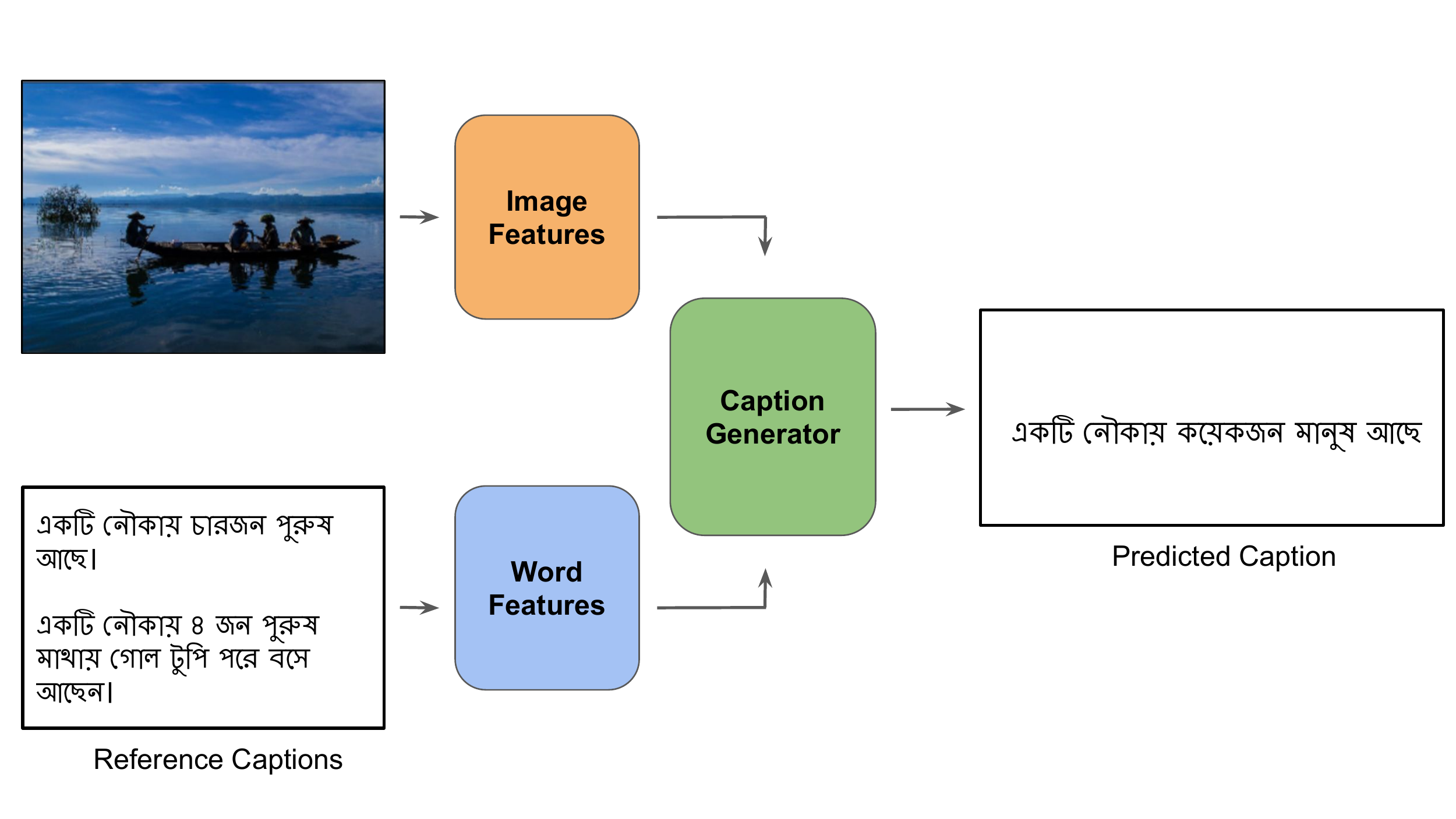}
\caption{Illustration of sequence-to-sequence basic architecture where image and linguistic features are merged to generate meaningful captions in the Bengali language.}
\label{figintro}
\end{figure}

The fields of computer vision and natural language processing have seen significant progress due to the recent development in deep learning. The task of image captioning lies at the intersection of these two. Most notable works are based on the encoder-decoder framework, which is very similar to the sequence-to-sequence model for machine translation \cite{mt1}. The framework contains a CNN-based image feature extractor and a recurrent neural network (typically LSTM \cite{lstm}) based caption decoder to generate the relevant words iteratively. The decoder's job is to take in the caption generated so far and predict the next word with the highest probability among all the vocabulary words until an ending token is generated. All the existing works in the Bengali language \cite{chittron,oboyob} follow the same architecture, as mentioned above. In \cite{role_of_rnn}, there is a comparison between the two architectures of image captioning. The first one is the \textbf{inject} architecture, where the RNN is used as a caption generator conditioned by the image features. The second one is the \textbf{merge} or \textbf{mixture} architecture, where the RNN is primarily used for encoding linguistic representation only. The encoded linguistic features and the image features are then merged and passed as input to a multi-modal layer that performs the word by word prediction of the caption. The comparative study shows that models with merge architecture serve better than models with inject architecture. In natural language processing tasks like chunking, part-of-speech tagging, and named entity recognition, CNN based models have provided faster and accurate results \cite{nlp_from_scratch}. Also, In \cite{cnn+cnn}, it has been shown that CNN-CNN models are competitive in performance with CNN-RNN models in terms of image captioning.

Inspired by the aforementioned successes of the fusion model and CNN in NLP, we propose an encoder-decoder-based model following merge architecture for Bengali image captioning. We used ResNet-50 \cite{resnet} for encoding image features and a one dimensional CNN for encoding the linguistic features. Unlike \cite{cnn+cnn}, the CNN used in our work is followed by a pooling layer for capturing meaningful and significant features. Later, both the image and language features are mingled and passed to the decoder to generate the image's caption. We evaluate our work on the BanglaLekhaImageCaptions dataset \cite{BanglalekhaImageCaptions}. The experimental results show that our model performs better than all the existing models in the Bengali language. We also conducted a qualitative and quantitative comparison between our CNN-CNN based mixture model and the CNN-LSTM based mixture model proposed in \cite{role_of_rnn}. The experimental results confirm that the CNN-based language encoder is responsible for the proposed model's overall better performance. In summary, the main technical contributions of this paper are the following:
\begin{itemize}
    \item We present CNN instead of regular LSTM to learn linguistic information and use it for word prediction during the caption decoding phase. Meanwhile, we use ResNet-50 \cite{resnet} architecture as an image feature extractor. 
    \item We qualitatively and quantitatively evaluate our approach on the BanglaLekhaImageCaptions dataset.
    \item Our model achieves SOTA performance on the BanglaLekhaImageCaptions dataset and outperforms the existing encoder-decoder models while describing complex scenes. We also present the human evaluated score for qualitative evaluation of the generated captions. The code is available on Github\footnote{https://github.com/FaiyazKhan11/Improved-Bengali-Image-Captioning-via-deep-convolutional-neural-network-based-encoder-decoder-model}
\end{itemize}

\section{Related Works}
In the early years of research in image captioning, many complex systems consisting of primitive visual object identifiers and language models were used \cite{trafic_image}. These systems were rule-based and predominantly hand-designed. Moreover, these systems worked only on a limited domain of images. In \cite{ob_mt}, image captioning was treated as a machine translation task. But this system failed to capture the fine-grained relationship among the objects in the image. Along with the advancements of deep learning methods, image captioning systems produced considerably improved performances following the same deep learning-based architecture as machine translation \cite{mt1,mt2}. These works adopted the same encoder-decoder framework \cite{show&tell,donahue,dense_cap} and framed the idea of image captioning as translating the image into text. These systems used CNN for encoding images and RNN for decoding the images into sentences. Later, attention mechanisms were introduced to mimic the human behavior of capturing only the important features in an image and translating them into a natural language description \cite{show_attend&tell,sem_attn}. These systems generated the captions conditioned by the attention at a specific place of the image at each time step. Most of the systems built for English language are evaluated on MSCOCO \cite{mscoco}, Flickr30k \cite{flickr30k} and Flickr8k \cite{flickr8k} datasets. 

Researches on other languages like Japanese \cite{japanese}, Chinese \cite{chinese}, German \cite{german}, Arabic \cite{arabic} etc have also been performed. Most of these research works are experimented and evaluated on the translated versions of the MSCOCO \cite{mscoco} and Flickr8k \cite{flickr8k} datasets in their respective languages. In \cite{chittron}, a Bengali image captioning dataset \cite{BanglalekhaImageCaptions} was introduced along with a model which is very similar to \cite{show&tell}. While the results generated are not accurate enough, but it surely instigated further research works in Bengali. In \cite{oboyob}, a comparative analysis of the existing encoder-decoder LSTM decoder based models was presented. The models were evaluated on a trimmed down, machine-translated version of the Flickr8k \cite{flickr8k} dataset in Bengali. The Bengali captions generated, however, do not maintain the typical Bengali sentence structure and lacks usability. In this work, we present an encoder-decoder model with a CNN language encoder. We also provide experimental results on the BanglaLekhaImageCaptions dataset comparing our work with the existing LSTM based models.

\section{Dataset}
We trained and evaluated our model on the BanglaLekhaImageCaptions dataset \cite{BanglalekhaImageCaptions}. It is a modified version from the one introduced in \cite{chittron}. It contains 9,154 images with two captions for each image. The captions are generated by two native Bengali speakers. While this data set is not big in volume compared to the existing datasets in the English language, it maintains relevance with the Bengali culture to some extent. But the dataset also has a considerable amount of human bias. This bias hinders any model's ability to describe non-human subjects. Also, the captions are not detailed in some cases, which causes the training and evaluation of any model to be not as accurate as expected.

To train our model, we divided the data set into three parts, which are train, test, and validation. For training, we used 7154 images. 1000 images were used during validation, and the rest 1000 images were used during testing.

\section{Model}
\begin{figure}
\includegraphics[width=\textwidth]{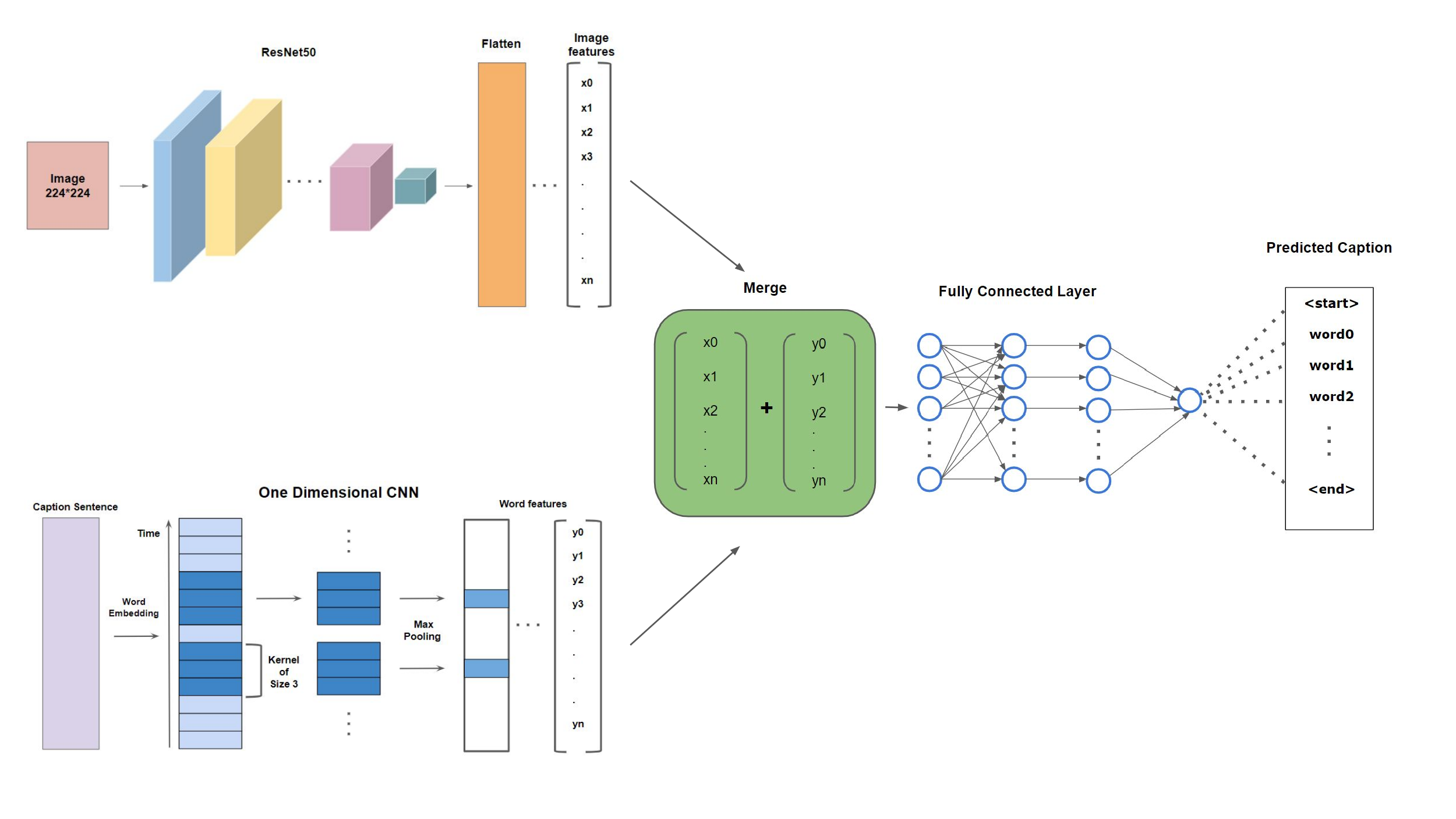}
\caption{The overview of the framework of our proposed CNN-ResNet-50 merged model, consisting of a ResNet-50 as image feature extractor and 1D-CNN with word embedding for generating linguistic information. Later, these two features are given as inputs to a multimodal layer that predicts what to generate next using this information at each time step}
\label{fig1}
\end{figure}

The model is based on encoder-decoder architecture. A two-dimensional convolutional neural network is used to encode the image features, and a one-dimensional convolutional neural network is used to encode the word sequences of the caption data. Later, both the encoded image and text features are merged and passed to a decoder to predict the caption in a word by word manner (figure \ref{fig1})

The model is divided into three parts:
\begin{itemize}
  \item \textbf{Image Feature Encoder:} We used pre-trained ResNet-50 \cite{resnet} as image feature extractor. It is trained on ImageNet  dataset \cite{imagenet}. Traditionally, neural networks with many layers tend to perform well in recognizing patterns. However, they also suffer from overfitting issues and are not easy to optimize. Residual CNNs are designed to have shortcut connections between layers. These connections perform identity mapping. ResNets are easy to optimize, and their performance increase with increasing network depth. We discard the final output layer of the ResNet-50 as it contains the output of image classification and used only the encoded image features produced by the hidden layers.
  \item \textbf{Word Sequence Encoder:}
 Two-dimensional convolutional neural networks have been extensively used in pattern recognition, image classification, and time series forecasting. The same property of these networks can be used in sequence processing. In our model, we used one-dimensional CNN for extracting one-dimensional patches from a sequence of words. The CNN has 512 filters with a kernel size of 3. The activation used is  Rectified Linear Units(ReLU). The CNN is followed by a Global Max Pooling Layer, which captures critical features from the convolutional layer's output.
	\item \textbf{Caption Generator:} The caption generator is a simple decoder containing a Dense 512 layer with ReLU activation. The output of the image feature encoder and word sequence encoder are combined by concatenation and used as input to the dense layer. The dense layer generates a softmax prediction for each word in the vocabulary to be the next word in the sequence, and the word with the highest probability is selected. This process continues until an ending token is generated.
\end{itemize}
The caption generator's output is transformed into a probability score for each word in the vocabulary. The greedy method chooses the word with the highest probability for each time step. This method may not always provide the best possible caption as any word's prediction depends on all the previously predicted words. So, it is more efficient to select the sequence with the highest overall score from a candidate of sequences. So we use the beam search technique with a beam size of 5. It considers the top 5 candidate words at the first decode step. For each of the first words, it generates five-second words and chooses the top five combinations of first and second words based on the additive score.  After the termination of five sequences, the sequence with the best overall score is selected. This method allows the process to be flexible and generate consistent results.

\section{Result and Analysis}
This section provides a quantitative and qualitative analysis of the performance of our model. While the evaluation metrics give a numeric idea about the captions' correctness, they can sometimes misinterpret any result. Qualitative analysis can evaluate the subtle difference in the generated captions compared with the natural human language description.

Predicted captions from the models were evaluated using the existing evaluation metrics \textbf{BLEU} \cite{bleu} (Bilingual Evaluation Understudy), \textbf{METEOR} \cite{meteor} (Metric for Evaluation of Translation with Explicit Ordering), \textbf{ROUGE} \cite{rouge} (Recall-Oriented Understudy for Gisting Evaluation), \textbf{CIDEr} \cite{cider} (Consensus-based Image Description Evaluation) and \textbf{SPICE} \cite{spice} (Semantic Propositional Image Caption Evaluation).

\begin{table*}[]
\centering
\renewcommand{\arraystretch}{2}
\caption{Quantitative analysis of performances among different models}
\resizebox{\textwidth}{!}{
\begin{tabular}{|c|c|c|c|c|c|c|c|c|c|}
\hline
\multicolumn{1}{|l|}{Search Type} &   \textbf{Models}             & BLEU-1 & BLEU-2 & BLEU-3 & BLEU-4 & METEOR & ROUGE-L & CIDEr & SPICE \\ \hline
\multirow{3}{*}{Greedy}           
& Our Model  & \textbf{0.651}  & \textbf{0.426}  & \textbf{0.278}   & \textbf{0.175}  & \textbf{0.297}  & \textbf{0.417}  & \textbf{0.501}  & \textbf{0.357}  \\ \cline{2-10}

& CNN + LSTM [mixture]\cite{role_of_rnn}         & 0.632  & 0.414  & 0.269  & 0.168  & 0.291  & 0.395   & 0.454 & 0.350 \\ \cline{2-10}

& CNN + Bi-LSTM [inject]\cite{chittron}    & 0.619  & 0.403  & 0.261  & 0.163  & 0.296  & 0.380   & 0.433 & 0.344 \\ \cline{2-10}

 & CNN + CNN [inject]        & 0.538
& 0.347  & 0.228  & 0.145  & 0.250  & 0.378   & 0.318 & 0.334 \\ \cline{2-10}
\hline

\multirow{3}{*}{Beam}
& Our Model        & \textbf{0.589}  & \textbf{0.395}  & \textbf{0.267}   & \textbf{0.175}  & \textbf{0.294}  & \textbf{0.434}  & \textbf{0.572}  & \textbf{0.353} \\ \cline{2-10}

& CNN + LSTM [mixture]\cite{role_of_rnn}    & 0.562  & 0.381  & 0.257  & 0.166  & 0.286  & 0.423   & 0.558 & 0.345 \\ \cline{2-10}

&  CNN + Bi-LSTM [inject]\cite{chittron}    & 0.575  & 0.374  & 0.241  & 0.149  & 0.286  & 0.412   & 0.532 & 0.349\\ \cline{2-10}

& CNN + CNN [inject]                        & 0.433  & 0.287  & 0.185  & 0.113  & 0.255  & 0.386   & 0.328 & 0.324 \\ \cline{2-10}
\hline
\end{tabular}
}
\label{table1}
\end{table*}

\begin{figure}[!htbp]
\centering
\begin{subfigure}[b]{\textwidth}
        \includegraphics[width=0.8\linewidth]{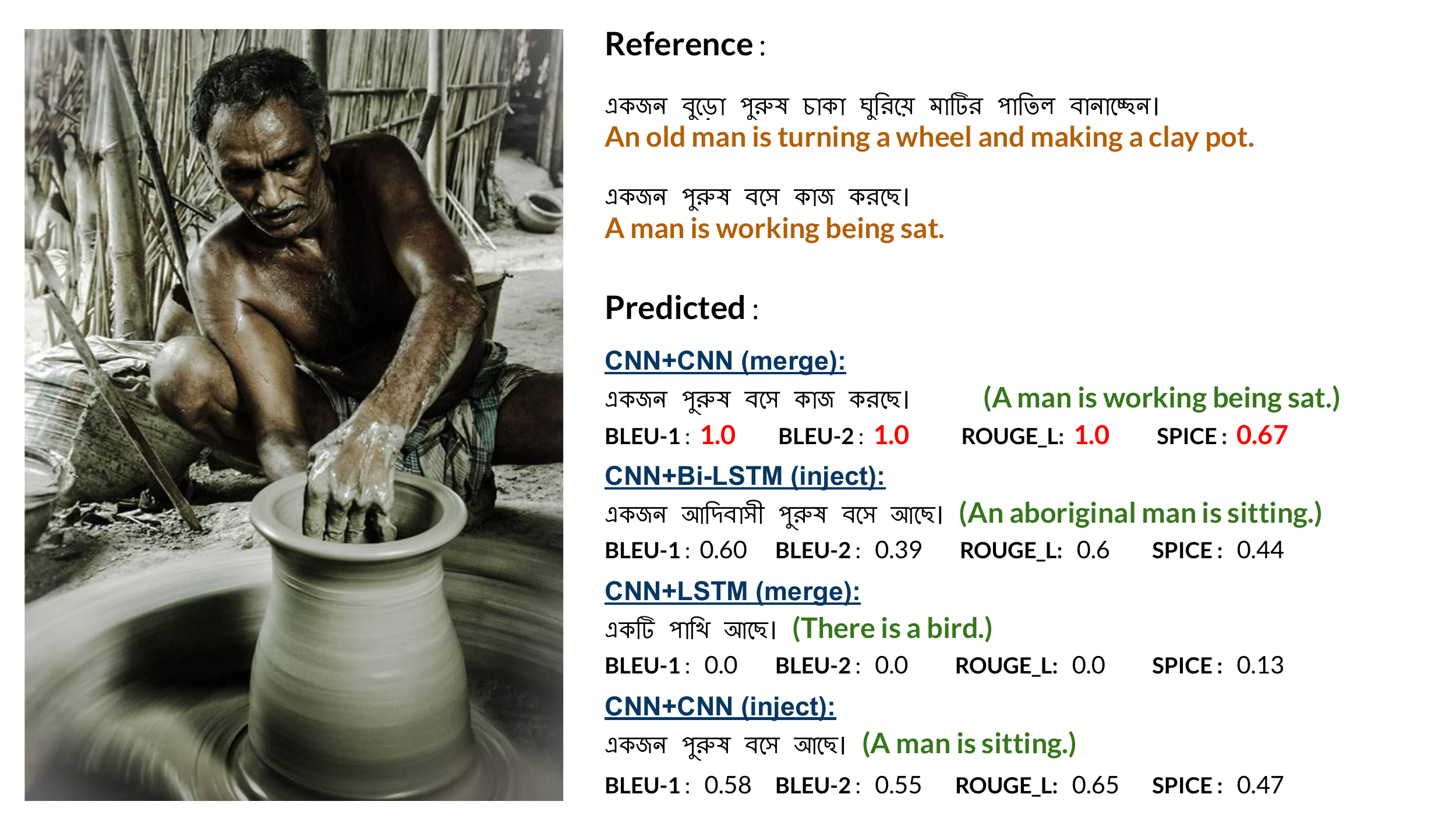}
         \caption{}
         \label{fig:manWorking}
     \end{subfigure}
     \hfill
     \begin{subfigure}[b]{\textwidth}
         \includegraphics[width=0.8\linewidth]{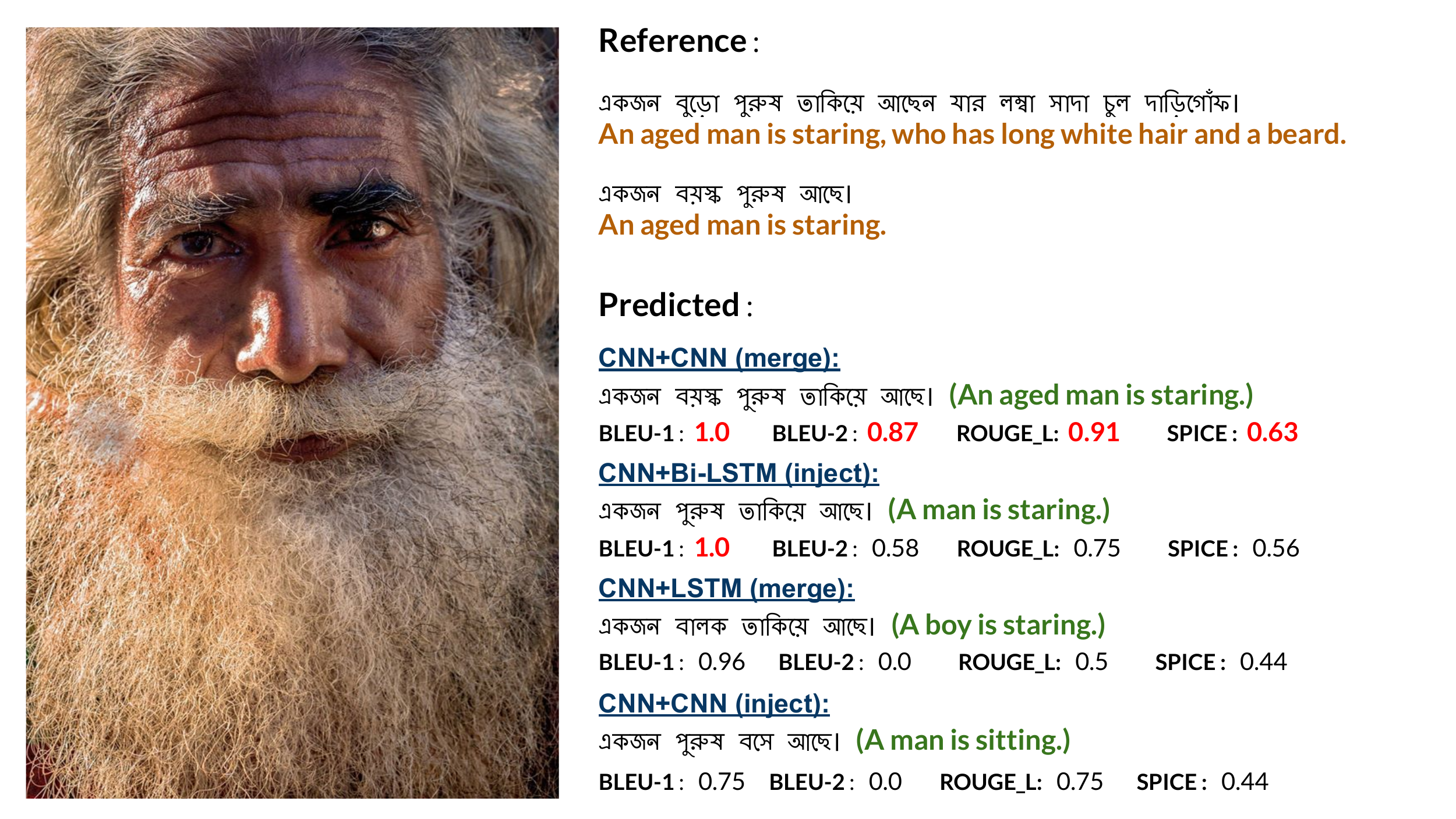}
         \caption{}
         \label{fig:oldMan}
     \end{subfigure}
        \caption{Comparison among the captions generated by different models}
        \label{comparison}
\end{figure}

A comparison among our model, inject architecture-based model with CNN language encoder, mixture architecture-based model with LSTM language encoder, and the model proposed in \cite{chittron} (Bi-directional LSTM language encoder with inject architecture) can be found in table \ref{table1}. We replicated the model of \cite{chittron} using the same ResNet-50 as image feature extractor instead of the VGG-16 \cite{vgg16} used in the original work to make sure that the better performance of our model is not only due to the better image model. We also present the scores of both the greedy and beam search method. From \ref{table1}, it can be seen, our model based on CNN word sequence processor has achieved better results in all the metrics than the traditional LSTM based models with both mixture and inject architectures and the CNN based models with inject architecture.  Our model's superior performance can be attributed to the one dimensional CNN model we used for sequence processing with the merge architecture. We used a window of size 3 for CNN. This window size with merge architecture enabled our model to learn words or word fragments of size 3. As a result, the fine-grained information present in the captions is learned during training. Following the CNN layer, the pooling layer filters only the significant features, which means the correlation between the words is stored better. Besides, this combination of one-dimensional CNN as a sequence processor with merge architecture can remember more diversified words while generating captions. These are evident in the comparison of the quality of the captions generated by the models in the figure--\ref{comparison}. The scores with the highest accuracy have been shown in table \ref{table1} with boldface. Among the evaluation metrics the most important metrics for evaluating image captions are \textbf{CIDEr} \cite{cider} and \textbf{SPICE} \cite{spice} since these are specially prepared for evaluating image captions. Better scores in these two metrics indicate the quality of performance of our model. The scores of the evaluation metrics were calculated using pycocoevalcap\footnote{https://github.com/salaniz/pycocoevalcap} library for python 3 available in github\footnote{https://github.com} which is a support for MS COCO caption evaluation tools \cite{mscoco_eval_tool}.

The performance comparison among our model and other models can be seen in figure--\ref{comparison}. We present the predicted Bengali caption and corresponding English Translated caption for non-native Bengali speakers. Our model performed better not only in the scores but also in the quality of captions. In figure--\ref{fig:manWorking}, it is observable that our model predicted the most relevant caption compared to other models describing the gender of the human subject and the work he is doing in the image.  In figure--\ref{fig:oldMan}, our model detected the gender correctly and captured the person's age range. It is also noticeable that all the metrics' score, including the SPICE, is better, which indicates the better quality caption.

\begin{figure}[!htbp]
\centering
\begin{subfigure}[b]{\textwidth}
        \includegraphics[width=.8\linewidth, height = 4.8cm]{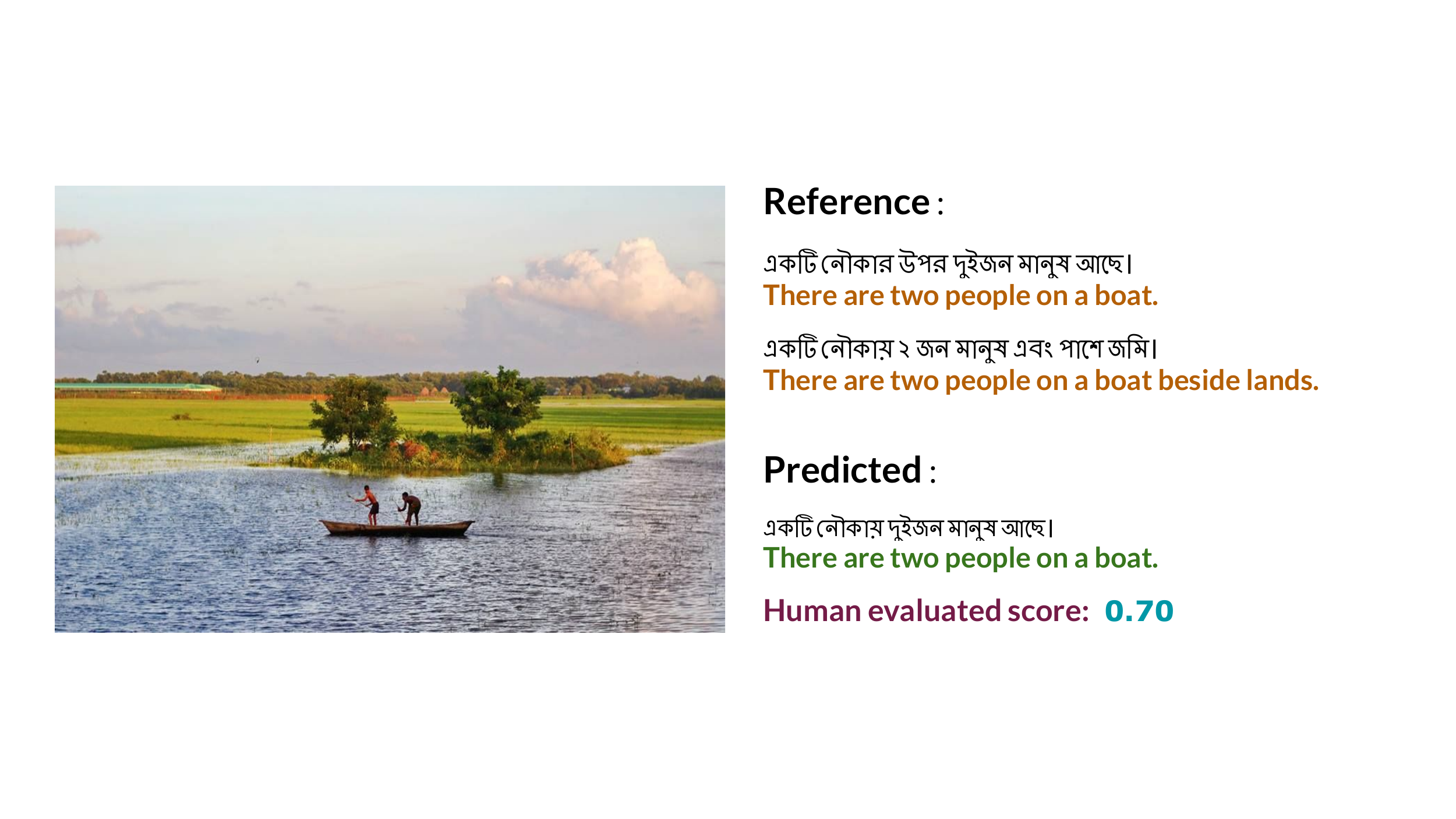}
         \caption{}
         \label{fig: 2per+boat}
     \end{subfigure}
     \hfill
     \begin{subfigure}[b]{\textwidth}
         \includegraphics[width=.8\linewidth, height = 4.8cm]{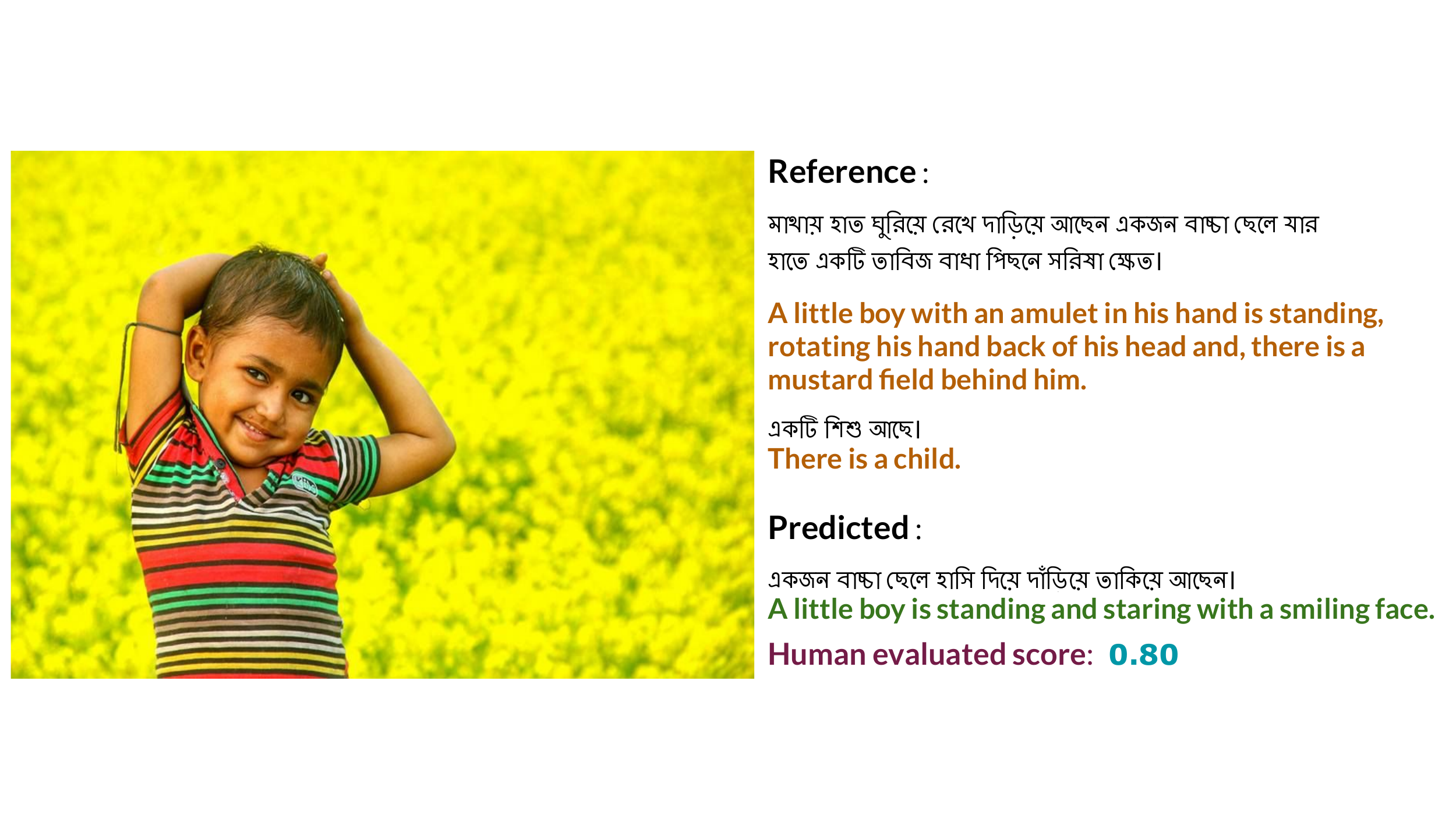}
         \caption{}
         \label{fig:child}
     \end{subfigure}
     \hfill
     \begin{subfigure}[b]{\textwidth}
         \includegraphics[width=.8\linewidth, height = 4.8cm]{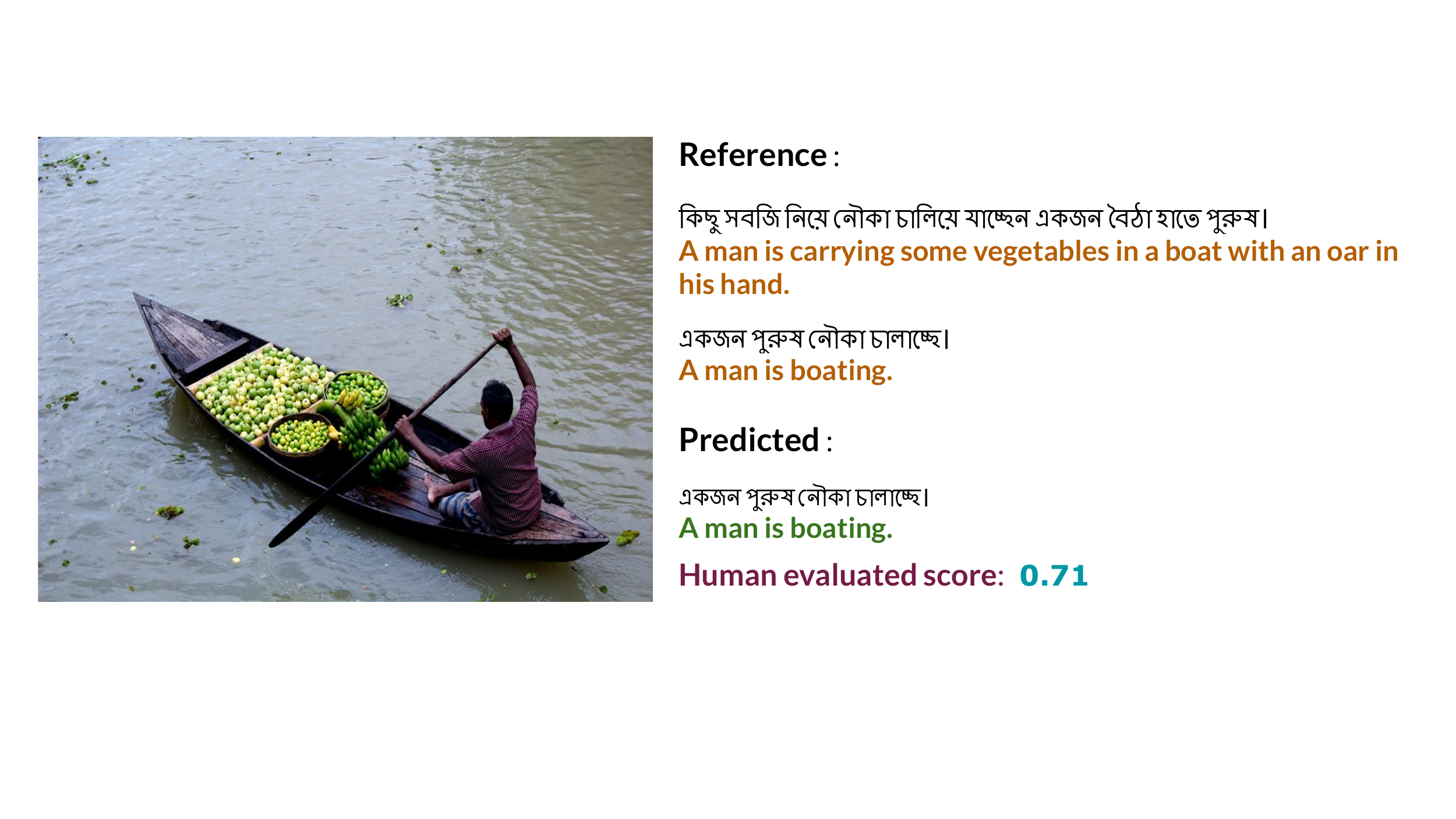}
         \caption{}
         \label{fig: 1per+boat}
     \end{subfigure}
     \hfill
     \begin{subfigure}[b]{\textwidth}
         \includegraphics[width=.8\linewidth, height = 4.8cm]{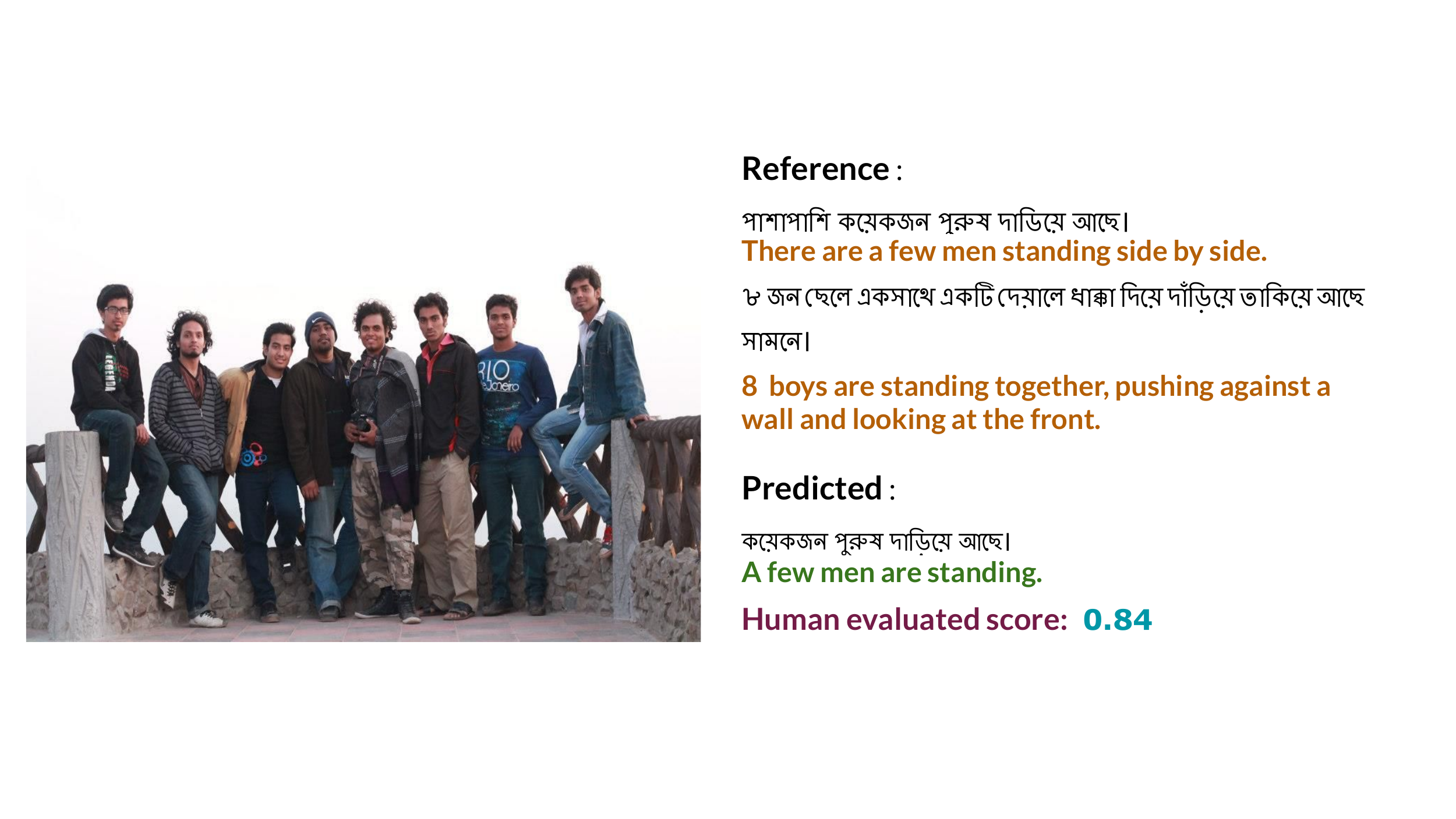}
         \caption{}
         \label{fig: peopleStanding}
     \end{subfigure}
        \caption{A glimpse of captions predicted by our model with qualitative evaluation}
        \label{samples}
\end{figure}

\subsection{Qualitative Analysis}
Some samples along with the English translation are depicted in the figure--\ref{samples} to represent the quality of captions generated. For qualitative evaluation, two native Bengali speakers were asked to give each caption a score between 0 and 1 from a set of sample images randomly selected from the test set. The scores are then averaged to generate the overall human evaluated score. In figure--\ref{fig: 2per+boat}, the model generated a decent caption correctly predicting the number of humans considering the human subjects' small size. In figure--\ref{fig:child}, our model described the expression of face and the person's age range, i.e., child correctly. The caption is also well detailed. Also, in figure--\ref{fig: 1per+boat} and figure--\ref{fig: peopleStanding}, the model described the content of the image in a very similar way to human description and hence, has achieved good evaluation scores.\begin{figure}[!htbp]
\centering
\begin{subfigure}[b]{\textwidth}
        \includegraphics[width=0.8\linewidth]{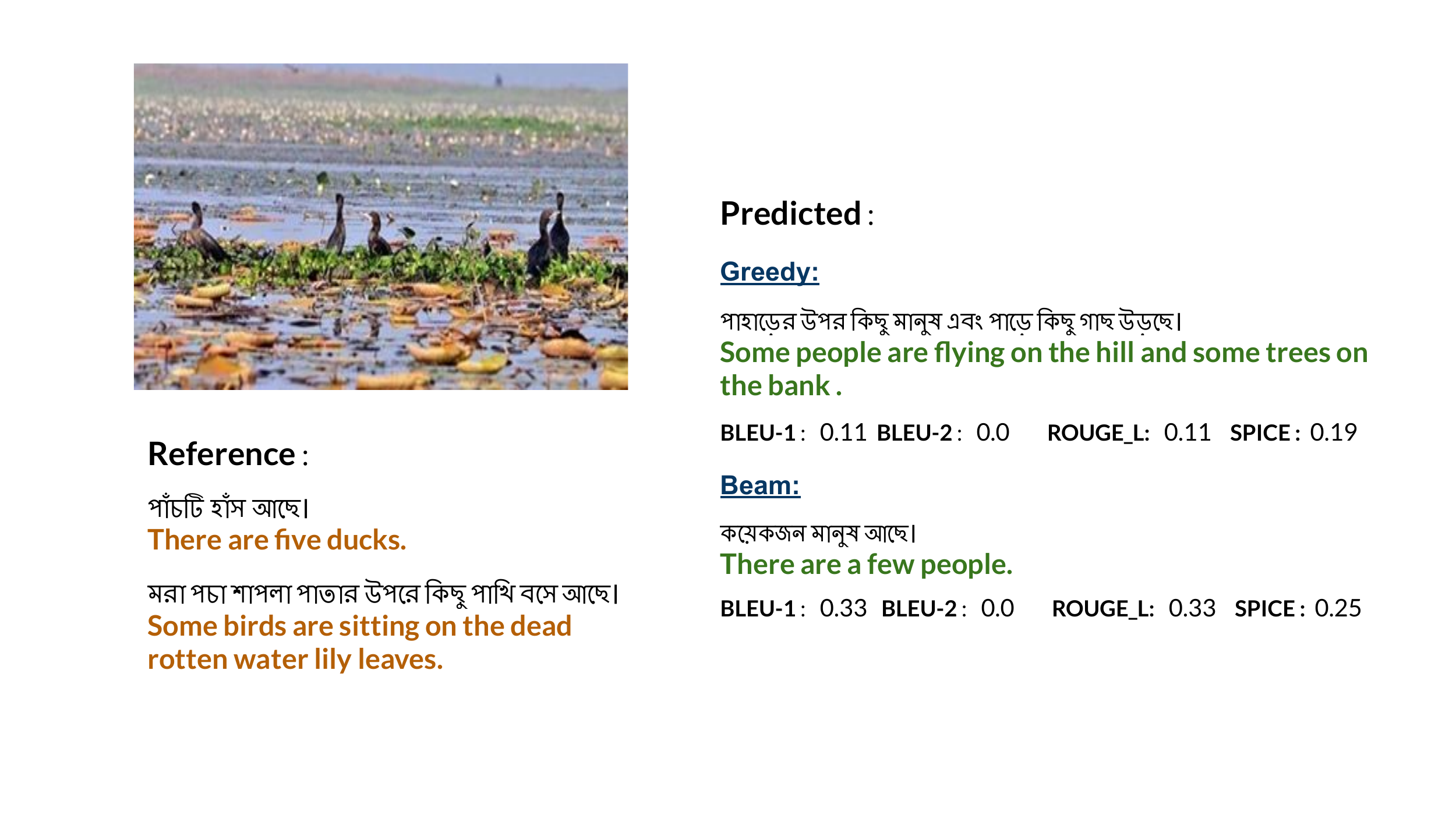}
         \caption{}
         \label{fig:bad1}
     \end{subfigure}
     \hfill
     \begin{subfigure}[b]{\textwidth}
          \includegraphics[width=0.8\linewidth]{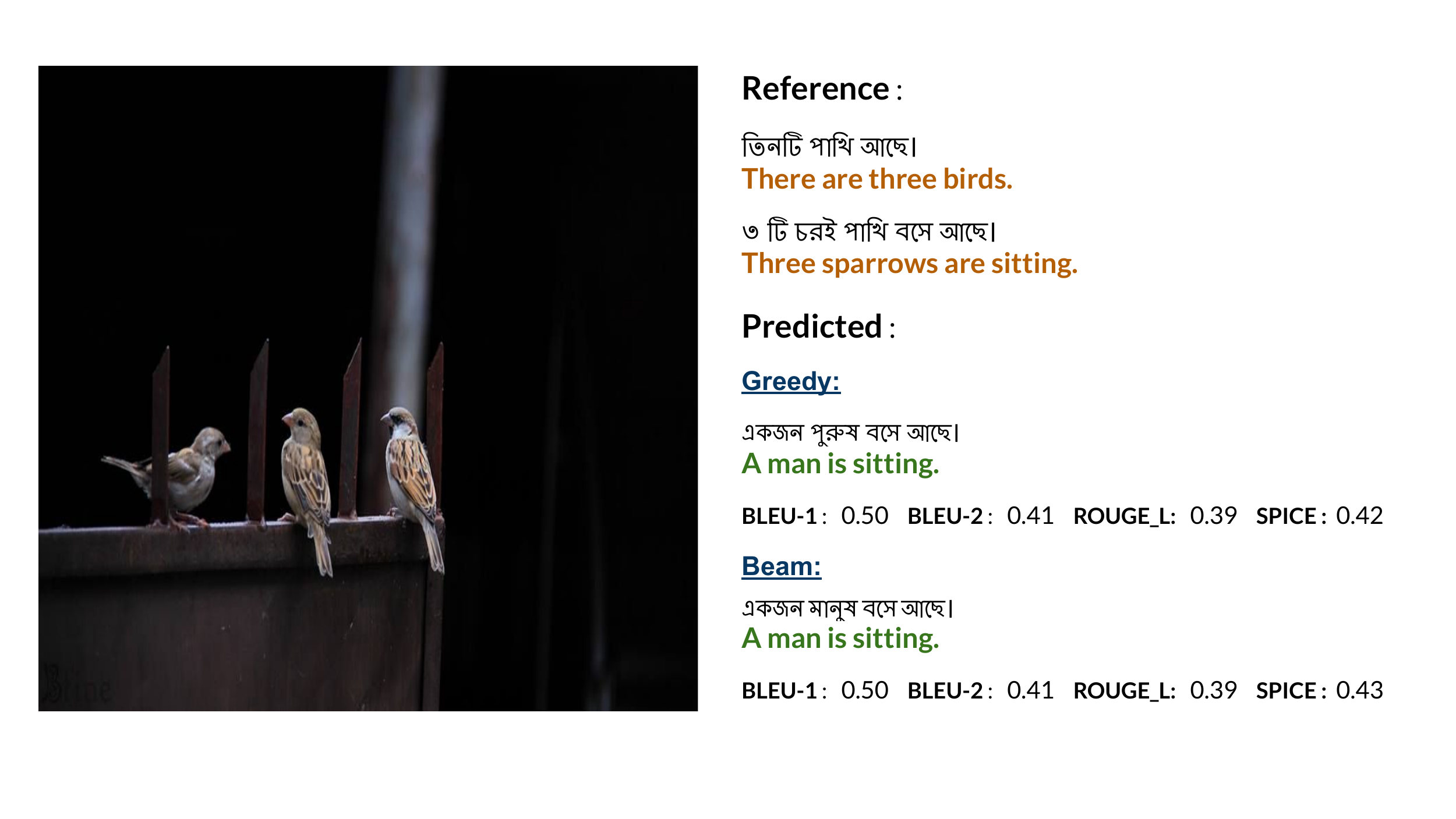}
         \caption{}
         \label{fig:bad2}
     \end{subfigure}
        \caption{Few incongruous captions generated by our model}
        \label{bad}
\end{figure}

In figure--\ref{bad}, some wrong predictions made by our model are shown. In figure--\ref{fig:bad1}, it could not correctly describe the content of the image at all, and in the figure--\ref{fig:bad2}, it misinterpreted the birds as persons. 

The faulty predictions made by our model can be attributed mainly to the dataset introduced by \cite{chittron}. In this dataset, most of the images contain human subjects which have almost similar types of captions. As a result, the model is trained with a massive amount of similar type human subjects and hence fails to detect and describe non-human subjects during testing. Besides, number of captions for each image is only two compared to five in the widely used English datasets. The dataset contains significant amount of spelling mistakes in Bengali caption which is also evident in the examples above. All of these are indicative of lacking details and variety in the dataset. 

\section{Conclusion}
This paper presents a CNN-CNN merged encoder-decoder-based image captioning system instead of a traditional sequence-to-sequence model. A substantial test conducted on the BanglaLekhaImageCaptions dataset with superlative performance validates the efficacy of our proposed model. Additionally, experimental results show that the CNN language model, combined with the merge architecture, captures the fined-grained sentence structure information with better linguistic diversity and produces more accurate and humanoid captions than the traditional LSTM. Nevertheless, the proposed model suffers from recognizing non-human subjects as the dataset is biased towards human subjects. This leaves us desired for a well-varied and detailed captioned dataset for Bengali Image Captioning. Therefore, we are motivated to develop a gold standard image caption dataset for Bengali for future work. Using other search methods such as constrained beam search in the decoding phase can be an area of future work. Besides, multilingual transformers available in NLP (Natural Language Processing) tasks like BERT \cite{bert}, XLM \cite{xlm}, XLNet \cite{xlnet} etc can generate promising results in the Bengali language.

\section{Acknowledgements}
We want to thank the Natural Language Processing Group, Dept. of CSE, SUST, for their valuable guidelines in our research work.

\bibliographystyle{splncs04}
\bibliography{main}

\begin{thebibliography}{10}
\providecommand{\url}[1]{\texttt{#1}}
\providecommand{\urlprefix}{URL }
\providecommand{\doi}[1]{https://doi.org/#1}

\bibitem{arabic}
Al-Muzaini, H.A., Al-Yahya, T.N., Benhidour, H.: Automatic arabic image
  captioning using rnn-lst m-based language model and cnn. International
  Journal of Advanced Computer Science and Applications  \textbf{9}(6) (2018)

\bibitem{spice}
Anderson, P., Fernando, B., Johnson, M., Gould, S.: Spice: Semantic
  propositional image caption evaluation. In: European Conference on Computer
  Vision. pp. 382--398. Springer (2016)

\bibitem{mt2}
Bahdanau, D., Cho, K., Bengio, Y.: Neural machine translation by jointly
  learning to align and translate. In: 3rd International Conference on Learning
  Representations, ICLR 2015 (2015)

\bibitem{mscoco_eval_tool}
Chen, X., Fang, H., Lin, T.Y., Vedantam, R., Gupta, S., Doll{\'a}r, P.,
  Zitnick, C.L.: Microsoft coco captions: Data collection and evaluation
  server. arXiv preprint arXiv:1504.00325  (2015)

\bibitem{nlp_from_scratch}
Collobert, R., Weston, J., Bottou, L., Karlen, M., Kavukcuoglu, K., Kuksa, P.:
  Natural language processing (almost) from scratch. Journal of machine
  learning research  \textbf{12}(ARTICLE),  2493--2537 (2011)

\bibitem{xlm}
Conneau, A., Lample, G.: Cross-lingual language model pretraining. In: Advances
  in Neural Information Processing Systems. pp. 7059--7069 (2019)

\bibitem{oboyob}
Deb, T., Ali, M.Z.A., Bhowmik, S., Firoze, A., Ahmed, S.S., Tahmeed, M.A.,
  Rahman, N., Rahman, R.M.: Oboyob: A sequential-semantic bengali image
  captioning engine. Journal of Intelligent \& Fuzzy Systems  \textbf{37}(6),
  7427--7439 (2019)

\bibitem{imagenet}
{Deng}, J., {Dong}, W., {Socher}, R., {Li}, L., {Kai Li}, {Li Fei-Fei}:
  Imagenet: A large-scale hierarchical image database. In: 2009 IEEE Conference
  on Computer Vision and Pattern Recognition. pp. 248--255 (2009).
  \doi{10.1109/CVPR.2009.5206848}

\bibitem{meteor}
Denkowski, M., Lavie, A.: Meteor universal: Language specific translation
  evaluation for any target language. In: Proceedings of the ninth workshop on
  statistical machine translation. pp. 376--380 (2014)

\bibitem{bert}
Devlin, J., Chang, M.W., Lee, K., Toutanova, K.: Bert: Pre-training of deep
  bidirectional transformers for language understanding. In: Proceedings of the
  2019 Conference of the North American Chapter of the Association for
  Computational Linguistics: Human Language Technologies, Volume 1 (Long and
  Short Papers). pp. 4171--4186 (2019)

\bibitem{donahue}
Donahue, J., Anne~Hendricks, L., Guadarrama, S., Rohrbach, M., Venugopalan, S.,
  Saenko, K., Darrell, T.: Long-term recurrent convolutional networks for
  visual recognition and description. In: Proceedings of the IEEE conference on
  computer vision and pattern recognition. pp. 2625--2634 (2015)

\bibitem{ob_mt}
Duygulu, P., Barnard, K., de~Freitas, J.F., Forsyth, D.A.: Object recognition
  as machine translation: Learning a lexicon for a fixed image vocabulary. In:
  European conference on computer vision. pp. 97--112. Springer (2002)

\bibitem{german}
Elliott, D., Frank, S., Sima’an, K., Specia, L.: Multi30k: Multilingual
  english-german image descriptions. In: Proceedings of the 5th Workshop on
  Vision and Language. pp. 70--74 (2016)

\bibitem{trafic_image}
Gerber, R., Nagel, N.H.: Knowledge representation for the generation of
  quantified natural language descriptions of vehicle traffic in image
  sequences. In: Proceedings of 3rd IEEE International Conference on Image
  Processing. vol.~2, pp. 805--808. IEEE (1996)

\bibitem{resnet}
He, K., Zhang, X., Ren, S., Sun, J.: Deep residual learning for image
  recognition. In: Proceedings of the IEEE conference on computer vision and
  pattern recognition. pp. 770--778 (2016)

\bibitem{lstm}
Hochreiter, S., Schmidhuber, J.: Long short-term memory. Neural computation
  \textbf{9}(8),  1735--1780 (1997)

\bibitem{flickr8k}
Hodosh, M., Young, P., Hockenmaier, J.: Framing image description as a ranking
  task: Data, models and evaluation metrics. Journal of Artificial Intelligence
  Research  \textbf{47},  853--899 (2013)

\bibitem{dense_cap}
Johnson, J., Karpathy, A., Fei-Fei, L.: Densecap: Fully convolutional
  localization networks for dense captioning. In: Proceedings of the IEEE
  conference on computer vision and pattern recognition. pp. 4565--4574 (2016)

\bibitem{chinese}
Li, X., Lan, W., Dong, J., Liu, H.: Adding chinese captions to images. In:
  Proceedings of the 2016 ACM on International Conference on Multimedia
  Retrieval. pp. 271--275 (2016)

\bibitem{rouge}
Lin, C.Y.: Rouge: A package for automatic evaluation of summaries. In: Text
  summarization branches out. pp. 74--81 (2004)

\bibitem{mscoco}
Lin, T.Y., Maire, M., Belongie, S., Hays, J., Perona, P., Ramanan, D.,
  Doll{\'a}r, P., Zitnick, C.L.: Microsoft coco: Common objects in context. In:
  European conference on computer vision. pp. 740--755. Springer (2014)

\bibitem{BanglalekhaImageCaptions}
Mansoor, N.K., Mohammed, A.H., Momen, N., Rahman, S., Matiur, M.:
  Banglalekhaimagecaptions, mendeley data  (2019),
  \url{http://dx.doi.org/10.17632/rxxch9vw59.2}

\bibitem{bleu}
Papineni, K., Roukos, S., Ward, T., Zhu, W.J.: Bleu: a method for automatic
  evaluation of machine translation. In: Proceedings of the 40th annual meeting
  of the Association for Computational Linguistics. pp. 311--318 (2002)

\bibitem{chittron}
Rahman, M., Mohammed, N., Mansoor, N., Momen, S.: Chittron: An automatic bangla
  image captioning system. Procedia Computer Science  \textbf{154},  636--642
  (2019)

\bibitem{vgg16}
Simonyan, K., Zisserman, A.: Very deep convolutional networks for large-scale
  image recognition. arXiv preprint arXiv:1409.1556  (2014)

\bibitem{mt1}
Sutskever, I., Vinyals, O., Le, Q.V.: Sequence to sequence learning with neural
  networks. In: Advances in neural information processing systems. pp.
  3104--3112 (2014)

\bibitem{role_of_rnn}
Tanti, M., Gatt, A., Camilleri, K.: What is the role of recurrent neural
  networks ({RNN}s) in an image caption generator? In: Proceedings of the 10th
  International Conference on Natural Language Generation. pp. 51--60.
  Association for Computational Linguistics, Santiago de Compostela, Spain (Sep
  2017)

\bibitem{cider}
Vedantam, R., Lawrence~Zitnick, C., Parikh, D.: Cider: Consensus-based image
  description evaluation. In: Proceedings of the IEEE conference on computer
  vision and pattern recognition. pp. 4566--4575 (2015)

\bibitem{show&tell}
Vinyals, O., Toshev, A., Bengio, S., Erhan, D.: Show and tell: A neural image
  caption generator. In: Proceedings of the IEEE conference on computer vision
  and pattern recognition. pp. 3156--3164 (2015)

\bibitem{cnn+cnn}
Wang, Q., Chan, A.B.: Cnn+ cnn: Convolutional decoders for image captioning.
  In: 31st IEEE/CVF Conference on Computer Vision and Pattern Recognition (CVPR
  2018) (2018)

\bibitem{show_attend&tell}
Xu, K., Ba, J., Kiros, R., Cho, K., Courville, A., Salakhudinov, R., Zemel, R.,
  Bengio, Y.: Show, attend and tell: Neural image caption generation with
  visual attention. In: International conference on machine learning. pp.
  2048--2057 (2015)

\bibitem{xlnet}
Yang, Z., Dai, Z., Yang, Y., Carbonell, J., Salakhutdinov, R.R., Le, Q.V.:
  Xlnet: Generalized autoregressive pretraining for language understanding. In:
  Advances in neural information processing systems. pp. 5753--5763 (2019)

\bibitem{japanese}
Yoshikawa, Y., Shigeto, Y., Takeuchi, A.: Stair captions: Constructing a
  large-scale japanese image caption dataset. In: Proceedings of the 55th
  Annual Meeting of the Association for Computational Linguistics (Volume 2:
  Short Papers). pp. 417--421 (2017)

\bibitem{sem_attn}
You, Q., Jin, H., Wang, Z., Fang, C., Luo, J.: Image captioning with semantic
  attention. In: Proceedings of the IEEE conference on computer vision and
  pattern recognition. pp. 4651--4659 (2016)

\bibitem{flickr30k}
Young, P., Lai, A., Hodosh, M., Hockenmaier, J.: From image descriptions to
  visual denotations: New similarity metrics for semantic inference over event
  descriptions. Transactions of the Association for Computational Linguistics
  \textbf{2},  67--78 (2014)

\end{thebibliography}

\end{document}